\documentclass[letterpaper, 10 pt, conference]{ieeeconf}  

\IEEEoverridecommandlockouts                              

\usepackage{amsmath}
\usepackage{amsthm,amsmath,amssymb}
\usepackage{mathrsfs}
\usepackage{bm}
\usepackage{algorithm}
\usepackage{algorithmic}
\usepackage{color}
\usepackage{graphicx}
\usepackage{makecell}
\usepackage{tabularx}
\usepackage{caption}
\usepackage{float} 
\usepackage{subcaption}
\usepackage{hyperref}

\title{\LARGE \bf
SpaceOctopus: An Octopus-inspired Motion Planning Framework for Multi-arm Space Robot
}

\author{Wenbo Zhao$^{1,*}$, Shengjie Wang$^{2,*}$, Yixuan Fan$^{1}$, Yang Gao$^{2}$, and Tao Zhang$^{1,\dag}$ \textit {Fellow, IEEE}
\thanks{Our website can be found at \href{https://sites.google.com/view/spaceoctopus}{https://sites.google.com/view/spaceoctopus}.}
\thanks{$^{*}$Equal contribution. $^{\dag}$Corresponding author: taozhang@tsinghua.edu.cn.}
\thanks{$^{1}$Department of Automation, Tsinghua University. $^{2}$Institute for Interdisciplinary Information Sciences, Tsinghua University.}%
\thanks{This research was funded by the National Natural Science Foundation of China under Grant U21B6002.}%
}

\begin{document}

\maketitle
\thispagestyle{empty}
\pagestyle{empty}

\begin{abstract}
Space robots have played a critical role in autonomous maintenance and space junk removal. 
Multi-arm space robots can efficiently complete the target capture and base reorientation tasks due to their flexibility and the collaborative capabilities between the arms. 
However, the complex coupling properties arising from both the multiple arms and the free-floating base present challenges to the motion planning problems of multi-arm space robots.
We observe that the octopus elegantly achieves similar goals when grabbing prey and escaping from danger.  
Inspired by the distributed control of octopuses' limbs, we develop a multi-level decentralized motion planning framework to manage the movement of different arms of space robots. 
This motion planning framework integrates naturally with the multi-agent reinforcement learning (MARL) paradigm. 
The results indicate that our method outperforms the previous method (centralized training). Leveraging the flexibility of the decentralized framework, we reassemble policies trained for different tasks, enabling the space robot to complete trajectory planning tasks while adjusting the base attitude without further learning. 
Furthermore, our experiments confirm the superior robustness of our method in the face of external disturbances, changing base masses, and even the failure of one arm.

\end{abstract}

\section{INTRODUCTION}

Space robots are usually applied to capture space debris, repair faulty satellites, and refuel other spacecraft to ensure the long-term stable operation of satellites and space stations \cite{jiang2022progress}. Beyond capturing debris and failed satellites, the maintenance and reorientation of the base of space robots are crucial for uninterrupted communication with the ground and for maintaining the optimal angle between the solar panel and the sun \cite{xu1993adaptive}. Since traditional base reorientation methods expend high-pressure gas from the satellite, which is a limited and valuable resource in outer space \cite{huang2015post}, the exploration of using robotic arms to adjust the posture of space robots has become a topic of interest. 
Recent work \cite{rudin2021cat} studies the attitude control of a base based on a quadruped robot with four 2-DoF limbs, but it does not extend to the more general field of space robots with multiple 6-DoF arms.

\begin{figure}[t]
\centering
\includegraphics[width=0.423\textwidth]{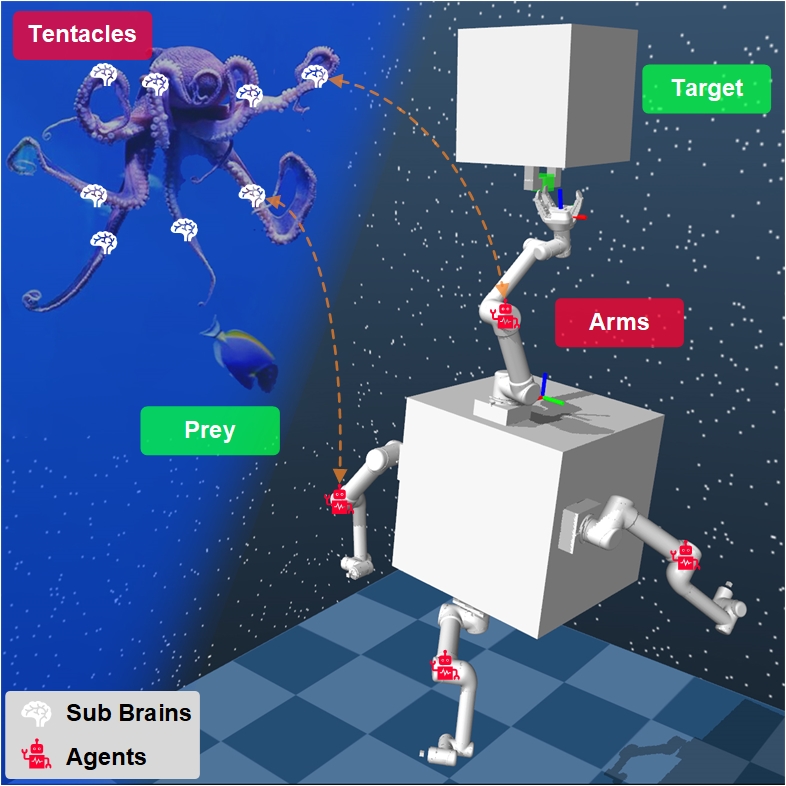}
\caption{The similarities between space robots and octopuses lie in their environments (zero-gravity outer space / underwater world), configurations (multiple robotic arms / multiple tentacles), and tasks (target capture / hunting). Inspired by the distributed brains of octopuses, we adopt a distributed control framework for space robots, in which each robotic arm learns its own strategy hierarchically for different tasks.}
\label{illu1}
\vspace{-15pt}
\end{figure}

Observing the similarities between the low-gravity conditions of outer space and the underwater environment, we seek inspiration from the hunting behaviors of intelligent underwater organisms for our control algorithm for the arms of space robots. The octopus, one of the most intelligent underwater creatures, has distributed brains, with each tentacle containing numerous neurons capable of independent thought \cite{hochner2012embodied}. This distributed control architecture improves their computational efficiency when hunting. As illustrated in Fig. \ref{illu1}, through coordination among its brains, an octopus can grasp prey with some tentacles while others adjust its position and posture—precisely the functionality desired for the target capture process of space robots. 

Inspired by octopuses, we design a hierarchical and distributed motion planning framework enabling the multi-arm space robot to perform diverse tasks including trajectory planning and base reorientation. 
The octopus-inspired planning method significantly reduces the difficulty of optimization by decomposing the original problem into multiple sub-problems.
First, the multi-level framework determines the sub-goal of each joint according to the whole task. 
Secondly, distributed learning trains the planning strategies to control each joint. 
Considering the complexity of model design and the singularity problem in traditional planning methods \cite{jiang2022progress, basmadji2020space,jia2017robust}, we adopt the model-free reinforcement learning (RL) method to obtain the planning strategies. Particularly, multi-agent RL makes our framework achieve decentralized learning of agents naturally.



Our contribution can be summarized as follows:

\begin{itemize}

\item Inspired by octopuses we devise a hierarchical and distributed framework for the motion planning problem of a multi-arm space robot.
In contrast to centralized learning, it alleviates the difficulty of optimization when decomposing into multiple sub-problems.

\item We design two basic tasks for multi-arm space robots, including trajectory planning and base reorientation. Due to the advantage of the MARL training paradigm, the trained policies outperform baseline methods in terms of precision and robustness. 

\item Leveraging the flexibility of decentralized control, we reassemble policies trained for different tasks onto the same space robot. Through coordination between agents, the space robot can complete the trajectory planning task while adjusting its base attitude without the need for further learning. 

\end{itemize}

\section{RELATED WORK}

The challenge of precisely controlling the base and robotic arms of space robots in low-gravity environments has been a subject of extensive study \cite{basmadji2020space,jia2017robust}. Umetani and Yoshida \cite{umetani1987continuous} developed an inverse kinematics solution using the Generalized Jacobian Matrix (GJM), although it was demonstrated in \cite{papadopoulos1993dynamic} that this solution encounters the singularity problem at some configurations. Efforts were also made in \cite{vafa1987dynamics}, where the Virtual Manipulator (VM) concept was introduced. However, a VM represents an idealized, massless kinematic chain, which cannot be physically built. Recently, researchers parameterize the joint trajectory using the sine polynomial function and formulate the planning problem as the non-linear optimization problem \cite{
chen2017path}. Nonetheless, most of these studies focus on trajectory planning with fixed initial and final points and require precise model of the kinematic and dynamic of the space robot. In terms of base attitude control, Kristiansen \textit{et al.} \cite{1470075} presented results on attitude control of a microsatellite by integrator backstepping, and Huang \textit{et al.} \cite{huang2015post} discussed adaptive control of a space robot system with an attitude-controlled base. Traditional methods consume precious gas, prompting authors in \cite{rudin2021cat} to utilize the limbs of a quadruped robot for attitude control, a concept partially aligned with our work.  However, their task is simpler, as the limb only has 2-DoF, whereas our work seeks to complete trajectory planning tasks while adjusting the base's attitude. 

Recently, reinforcement learning (RL) based methods have proven effective and robust in robotic control tasks \cite{ju2022transferring} and have been applied to space robot control tasks \cite{wu2020reinforcement, jia2016trajectory, yan2018control}. For example, Du \textit{et al.} \cite{du2019learning} proposed a controller for a free-floating space robot to capture targets without the kinematic and dynamic equations. However, these methods mainly focus on single-target trajectory planning, and often fail when the target position changes. Wang \textit{et al.} presented a multi-target trajectory planning method via reinforcement learning \cite{wang2021multi, wang2021end}. The aforementioned works widely demonstrate the effectiveness and robustness of model-free reinforcement learning algorithms for solving motion planning problems of space robots. Nevertheless, as the number of robotic arms increases, the exploration space of the agent grows exponentially, and to our knowledge, there has yet to be a successful application of RL-based methods on space robots with more than two arms. To address the large exploration space issue, a decentralized motor skill learning paradigm was proposed in \cite{guo2023decentralized}, where different sets of joints are designed to learn distinct policies for better robustness and generalization. However, without a centralized training with decentralized execution paradigm, it is challenging for the agents to coordinate their actions.

\section{PRELIMINARY}

\subsection{Simulation Environment}


\begin{figure}[t]
    \begin{subfigure}[t]{0.24\textwidth}
           \centering
           \includegraphics[width=\textwidth]{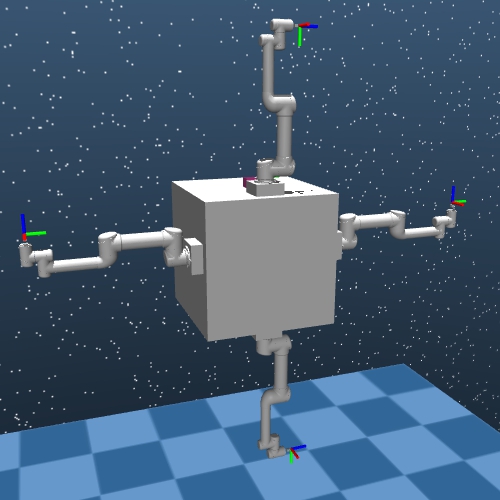}
            \caption{}
            \label{fig:Case1}
    \end{subfigure}
    \begin{subfigure}[t]{0.24\textwidth}
            \centering
            \includegraphics[width=\textwidth]{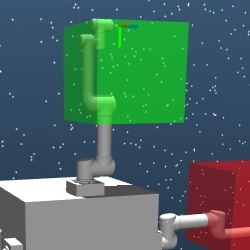}
            \caption{}
            \label{fig:b}
    \end{subfigure}
    \caption{(a) Simulation environment of the four-arm space robot. (b) In the trajectory planning task, the target position is sampled in a 0.3 $\times$ 0.3 $\times$ 0.3 $\mathrm{m}^3$ cube in front of each arm with randomly sampled desired orientation.}
\end{figure}

We utilize the dynamic simulation environment, MuJoCo \cite{todorov2012mujoco}, to construct the environment of a four-arm free-floating space robot, as illustrated in Fig. \ref{fig:Case1}. Four 6-degree-of-freedom (6-DoF) UR5 robotic arms are rigidly attached to the base of the space robot, with parameters identical to those of the actual robot. In the trajectory planning task, the goal for each end-effector is to reach a target randomly selected from an area within a 0.3 $\times$ 0.3 $\times$ 0.3 $\mathrm{m}^3$ cube positioned in front of each arm, along with a randomly sampled desired orientation. For the base reorientation task, the desired base attitude is randomly determined, ranging from $-$0.2 rad to 0.2 rad along every axis. The mass of the base is 400 kg with its size  0.8726 $\times$ 0.8726 $\times$ 0.8726 $\mathrm{m}^3$. Assuming the gripper of the robotic arm is insensitive to the shape of the object, we disregard the shape of grippers. Additionally, we omit the modeling of solar panels due to their negligible impact on the base, and the entire system is unaffected by gravity.

\subsection{Multi-Agent Reinforcement Learning}

To enable a distributed control strategy, we model the control procedure of the multi-arm space robot as a Decentralized Partially-Observable Markov Decision Process (DEC-POMDP), consisting of a tuple $G = \langle S, U, P, R, Z, n, O, \gamma\rangle$. Here, $s \in S$ represents the true state of the environment, and $U$ denotes the individual action space. At each time step, each agent $a \in A \equiv \{1,...,n\}$ selects an individual action $u^a \in U$, forming a joint action $\mathrm{\textbf{u}} \in \mathrm{\textbf{U}} \equiv U^n$, where $n$ indicates the number of agents. The state transition function $P(s'|s,\mathrm{\textbf{u}}): S\times\mathrm{\textbf{U}}\times S \to [0,1]$ determines the next state of the environment, and each agent gets an individual reward $R^a(s,\mathrm{\textbf{u}}):S\times\mathrm{\textbf{U}}\to \mathbb{R}$ based on the global state and their joint action. Note that in the base reorientation task the agents share the same reward, whereas in the trajectory planning task different agents are assigned diverse rewards. The partial observability is described by the observation function $O(s,a): S\times A\to Z$, where $Z$ is the individual observation space. In most MARL settings, the policy function and value function of each agent depends on an action-observation history $\tau^a\in T\equiv (Z\times U)^*$, and use RNN \cite{elman1990finding} instead of MLP as the policy network. However, we find that in robot control tasks, providing the agent with real-time joint angle and velocity observations renders the use of RNN layers unnecessary. Therefore, we employ an MLP for the structure of the policy and value networks.

\section{Methodology}

In this section, we formulate the trajectory planning and base reorientation problems of the multi-arm space robot. Inspired by the hierarchical and decentralized movement of octopus tentacles, we model the optimization problems as multi-agent RL problems. To effectively partition the tasks among multiple agents, we establish a three-level structure comprising the single-arm level, multi-arm level, and task level. By leveraging the hierarchical and decentralized framework, we are able to simplify the optimization process and enhance scalability and robustness for multiple robotic arms.



\subsection{Key Insight from Octopus}


We propose a method to mitigate the training challenges posed by the high dimensionality of the action space and the unclear rewards in controlling multi-arm space robots. Our inspiration comes from octopuses, which possess not only a central brain but also numerous neurons distributed across each tentacle, enabling independent thought and the execution of specific tasks.  This distribution of "brains" simplifies the learning process, enhances reaction speed and efficiency, and allows for the simultaneous completion of diverse tasks by different tentacles.
Taking inspiration from this, we transform the control problem of multi-arm space robots into a multi-agent reinforcement learning problem, wherein different arms learn and train under the CTDE \cite{lowe2020multiagent} paradigm. We further utilize the structural characteristics of the robotic arm to divide the learning objectives of the agent. Through distributed training, we significantly reduce the action space for each agent, decrease exploration complexity, and provide more direct reward guidance to individual agents, thereby increasing learning speed and efficiency. Similarly to octopuses, distributed training enables each agent to operate independently and complete tasks using varied strategies without the need for retraining.

\subsection{Problem Formulation}
\label{pro_for}
As the trajectory planning and base reorientation problems are modeled as DEC-POMDP, each distributed controller is implemented by a policy network, which consists of a multi-layer perceptron (MLP) with two hidden layers and a driver layer. The driver layer is realized by a PD controller with fixed parameters for velocity tracking, and the reinforcement learning algorithm updates only the parameters of the policy layer. At each time step during the control process, each policy layer first maps the current individual observation to the expected velocities of specific motor joints, with the driver layer then maps these desired velocities to joint torques $\tau$. The torques are finally applied to control the four-arm space robot in simulation.

\subsubsection{Trajectory Planning}

The goal of trajectory planning is to minimize the position and orientation error between each end-effector and its target, defined as: $e_p^i = p_d^i - p^i$ and $e_\phi^i = \phi^i_d - \phi^i$
respectively, where $p_d^i$ and $\phi^i_d$ denotes the desired position and Euler angle of the end-effector, and $i=\{1, 2,3,4\}$ represents the arm index. Moreover, to ensure smooth control of the arms and reduce energy consumption during the capture process, the reward function is generalized as $r_t = \mathcal{R}(e_p, e_\phi, a_t. a_{t-1}, \tau_t)$, where $a_t$ and $a_{t-1}$ stands for the joint output of policy networks at current and last time step, and $\tau$ for the output of the driver layer. $e_p$ and $e_\phi$ represent the position and orientation error for all arms. To avoid collision and constraint the maximum value of joint motion, the mathematical description of optimization in trajectory planning is defined in Eq. 
\eqref{TPeq}: 
\begin{equation}
\begin{split} 
\label{TPeq}
&\max_{a\sim\pi} \quad \mathcal{J}=\int_{t=0}^T[r_t] \, dt\\
&s.t.\quad  \left\{ \begin{array}{lc} r_t = \mathcal{R}(e_p, e_\phi, a_t. a_{t-1}, \tau_t) \\
\Vert {q_t} \Vert \leq U_{q}\\
\Vert \dot{q_t} \Vert \leq U_{\dot{q}}\\
\Vert {\tau_t} \Vert \leq U_{\tau}\\
S\cap o =\varnothing \\
\end{array}\right.
\end{split}
\end{equation}
where $T$ denotes the maximum episode length and $U_*$ is the upper bounds of the joint parameters. Furthermore, $S$ represents the space occupied by the space robot, $o$ represents the space occupied by other objects, indicating the robot's capability to avoid collisions.

\subsubsection{Base Reorientation}

\begin{figure*}[t]
\centering
\includegraphics[width=0.96\textwidth]{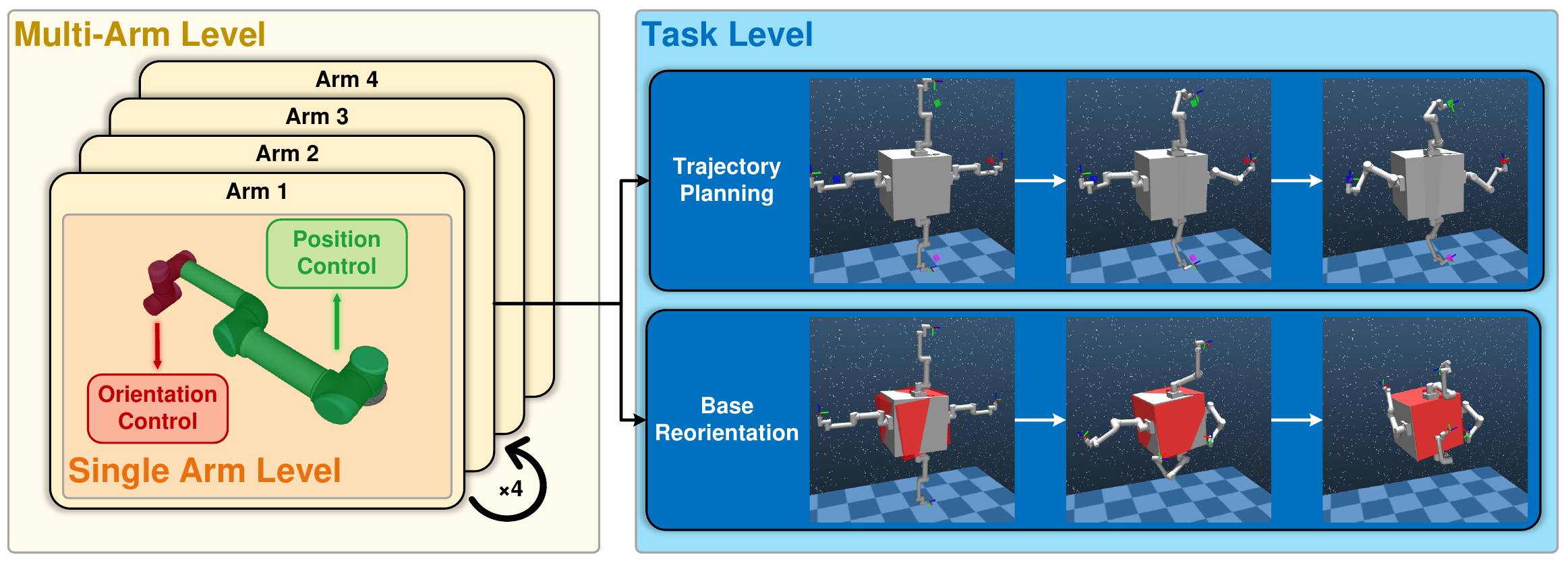}
\caption{The agent division methodology comprises three levels: the single-arm level, the multi-arm level, and the task level. To enhance the capabilities of the space robot, at the task level we can assign various tasks, such as trajectory planning and base reorientation, to the multiple arms of the space robot. The six joints of each arm are controlled by two controllers to achieve the desired position and orientation of the end-effector respectively.}
\label{illustration}
\end{figure*}

Unlike the trajectory planning task, the goal of the base reorientation task is to minimize the error between the desired and current attitude: $e_\phi ^b = \phi^b_d-\phi^b$, where the superscript $b$ denotes the base of the space robot, and the subscript $d$ denotes the desired attitude. Note that in this work we represent the attitude of the base or the orientation of the end-effector using Euler angles. Similarly, we can write out the mathematical description of the base attitude adjustment task: 
\begin{equation}
\begin{split} 
\label{BAAeq}
&\max_{a\sim\pi} \quad \mathcal{J}=\int_{t=0}^T[r_t] \, dt\\
&s.t.\quad  \left\{ \begin{array}{lc} r_t = \mathcal{R}(e_\phi^b, a_t. a_{t-1}, \tau_t) \\
\Vert {q_t} \Vert \leq U_{q}\\
\Vert \dot{q_t} \Vert \leq U_{\dot{q}}\\
\Vert {\tau_t} \Vert \leq U_{\tau}\\
S\cap o =\varnothing \\
\end{array}\right.
\end{split}
\end{equation}

\subsection{Division of Agents}

To enable distributed control, we hierarchically divide the motor joints of the multi-arm space robot into different groups and leave them to different agents to control. The division of agents is mainly in accordance with three levels, which is illustrard in Fig. \ref{illustration}.

\subsubsection{Single Arm Level}

We assign the motor joints of a single robotic arm to two individual controllers, viewing them as two distinct agents with diverse optimization goals. A key insight is that most robotic arms exhibit a clear structural characteristic: the first few links connected to the base are usually long, providing a larger operation space for the arm, while latter links are relatively short to enable agile operation of the end-effector. 

Leveraging this structural feature, we divide the motor joints on a single robotic arm into two sets, each controlled by a separate agent. Specifically, for the UR5 robotic arm used in our simulation environment, we assign the first three joints connected to the base
to the first agent and the remaining three wrist joints to the second. The first agent focuses on learning how to control the end-effector to reach the target position, while the second concentrates on achieving the desired orientation. Consequently, the reward function in Eq. \ref{TPeq} can be rewritten as: $r_p = \mathcal{R}(e_p,a_t, a_{t-1},\tau_t)$ and $r_\phi = \mathcal{R}(e_\phi,a_t, a_{t-1},\tau_t)$ for a single arm's position and orientation, respectively. It is important to note that the two agents do not share the same reward, and since any individual action of the joints will influence both the position and orientation of the end-effector, the agents on the same arm musr learn to cooperate to achieve high rewards of both.  

\subsubsection{Multi-Arm Level}

The robotic arms on the four-arm space robot are mounted in different positions and are in different postures. Consequently, even if the arms are identical, the same actions at the joints will result in different outcomes for the end-effector, depending on its installation pose. To this end, we equip the four arms with distinct control strategies, and for each arm we further divide the joints into two sets using the division method described in the previous section. As a result, the four-arm space robot system comprises a total of eight agents, with each agent controlling three motor joints. The reward function can be rewritten as $r_p^i = \mathcal{R}(e_p^i,a_t^i, a_{t-1}^i,\tau_t^i)$ and $r_\phi^i = \mathcal{R}(e_\phi^i,a_t^i, a_{t-1}^i,\tau_t^i)$, where $i \in \{1, 2, 3, 4\}$ denotes the arm index. While it may seem that the individual reward is solely related to the controlled joints, the action of any joint can affect the base's position and orientation, thereby influencing the end-effector error of each agent. Thus, the two agents on the same arm have to cooperate to achieve high position and orientation precision, while agents on different arms also have to coordinate their actions as the movement of one arm can affect the base and, consequently, the performance of the other arms.

\subsubsection{Task Level}
The aforementioned division methods are primarily designed for the trajectory planning task, in which the arms are controlled to reach the desired position and orientation of the end-effector.
For the base reorientation task, we similarly assign the control task of the four arms to eight agents, considering the structural characteristics of a single arm and the distinct poses of the four arms. 

Our key insight is that different sets of links also have varying impact on the base attitude: the first few links attached to the base being long and heavy with large inertia, significantly influence the base's attitude over a long duration and are suitable for broad-range attitude adjustments. In contrast, the shorter wrist links, with their smaller inertia, are ideal for fine-tuning the base's attitude. 
Besides, the task-level division also offers another benefit: it allows different arms to perform different tasks such as reaching targets or the reorientation of the base. It extremely improves the execution of mixed tasks by offering different tasks for multiple arms. 
Therefore, despite the difference in tasks we adopt the eight-agent division paradigm, while we train different policies to complete trajectory planning and base reorientation tasks. 

\section{OPTIMIZATION ALGORITHM AND \\TRAINING DETAILS}

In this section, we describe the training algorithm for the aforementioned MARL problem. 

\subsection{Optimization Algorithm}

The goal of MARL is to find an optimal joint policy $\bm{\pi}^*=\{ \pi^{1*},...,\pi^{n*}\}$. Here we denote the joint policy $\bm{\pi} = \{\pi^1,...,\pi^n\}$ and the discount sum of return $R^a_t=\sum_{k=0}^{\infty} \gamma^k r_{t+k}^a$, and we use the following standard definitions of the individual state-action value function $Q^{\bm{\pi}}$, the state value function $V^{\bm{\pi}}$ , and the advantage function $A^{\bm{\pi}}$: 
\begin{equation}
\begin{aligned}
\label{totaldef}
&Q^{\bm{\pi}}_a(s_t,\mathrm{\textbf{u}_t}) = \mathbb{E}_{s_{t+1}, \mathrm{\textbf{u}_{t+1}},...} [R^a_t|s_t,\mathrm{\textbf{u}_t}] \\
&V^{\bm{\pi}}_a(s_t) = \mathbb{E}_{\mathrm{\textbf{u}_t, s_{t+1}, ...}} [R^a_t|s_t] \\
&A^{\bm{\pi}}_a(s_t,\mathrm{\textbf{u}_t}) =Q^{\bm{\pi}}_a(s_t,\mathrm{\textbf{u}_t}) - V^{\bm{\pi}}_a(s_t) \\
&u_t^a\sim\pi^a(\cdot|s_t), s_{t+1}\sim P(\cdot|s_t,\mathrm{\textbf{u}_t})
\end{aligned}
\end{equation}
Note that for tasks that all agents share the same reward, the discount sum of reward $R^a_t$ remains unchanged for all agents, and the aforementioned equations should also be the same among the agents. The objective of the MARL problem is to find an optimal joint policy $\bm{\pi}^*$ that maximize the value of the initial state:
\begin{equation}
    \bm{\pi}^*=\mathop{\arg\max}\limits_{\bm{\pi}} \mathbb{E}_{s_0\sim\rho_0}\{V^{\bm{\pi}}_a(s_0)\}
\end{equation}
where $\rho_0$ is the distribution of the initial state. Note that under the base reorientation scenario where agents share the same reward, the agents will optimize the same function $V^{\bm{\pi}}$. However, for reward-unshared tasks such as trajectory planning, the distributed agents optimize different functions $V^{\bm{\pi}}_a$, which is dependent on others' policies. 

To train the individual agents to learn the optimal policy, we adopt the Centralized Training with Decentralized Execution (CTDE) structure \cite{oliehoek2016concise}, and use Multi-Agent Proximal Policy Optimization (MAPPO) \cite{guo2020joint, yu2022surprising} as the training algotithm. In detail, for each agent $a$ we use an individual neural network to parameterize its policy $\pi^a$ with $\theta_a$, which takes the current observation as input and outputs the action of joints under control. Meanwhile, we device a critic network $V^a(s)$ for each agent with parameters $\omega_a$, which takes the global state as input and outputs the state value. For simplicity, we denote the actor and critic for agent $a$ as $\pi_\theta^a(\cdot|z^a)$ and $V_\omega^a(s)$, respectively. The critic is updated to minimize the TD error \cite{sutton2018reinforcement}: 

\begin{equation}
\label{TDerror}
    J(\omega_a)=\mathbb{E}[V_\omega^a(s_t)-y_t^a]
\end{equation}
where $y_t^a=r_t^a+\gamma V_{\bar{\omega}}^a(s_{t+1})$, and $V_{\bar{\omega}}$ is the target state-value function with the parameters $\bar{\omega}_a$ periodically updated with the most recent $\omega_a$ to stabilize learning process. Note that the critic takes true state $s$ instead of individual observation $z^a$ as input, thus can gather global knowledge and other agents' information to guide the training process of the actor. The actor of each agent is updated according to the clipped policy gradient: 

\begin{equation}
\label{MAPPOEQ}
    \max\limits_{\theta_a} \mathbb{E}_{\bm{\pi}_{old}} \Big\{ \min \big\{ r(\theta_a)A^{\bm{\pi}_{old}}_a(s,\mathrm{\textbf{u}}), c^\epsilon (r(\theta_a)) A^{\bm{\pi}_{old}}_a(s,\mathrm{\textbf{u}}) \big\} \Big\}
\end{equation}
where the clip function $c^\epsilon(x)$ restricts $x$ into the interval $[1-\epsilon,1+\epsilon]$, and $r(\theta_a) = \dfrac{\pi^a(u^a|z^a)}{\pi^a_{old}(u^a|z^a)} $ is the probability ratio. The advantage term in Eq. \eqref{MAPPOEQ} is estimated by General Advantage Estimation (GAE) \cite{schulman2015high} with the learned state value function in Eq. \eqref{TDerror}.

\subsection{Training Details}

\subsubsection{State And Action}

As mentioned earlier, we divide the joints of the four-arm space robot into eight agents despite the different tasks of trajectory planning or base orientation. For the agents who control the first three links of the arm, their observation consists of a 28-dimensional vector and can be written as Eq. \eqref{OBfun1}:
\begin{equation}
\label{OBfun1}
    o_t^a=\langle p_b,\phi_b,v_b,\omega_b,q_p^a,\dot{q_p^a},p_e,p_d\rangle
\end{equation}
where $p_b$ and $\phi_b$ stand for the current position and orientation of the base, $v_b,\omega_b$ for the corresponding velocity and angular velocity.  $q_p^a$ and $\dot{q_p^a}$ denote the joint angle and angular velocity under control, and the current position of the end-effector and the corresponding desired position is denoted by $p_e$ and $p_d$ respectively. $a = \{1,3,5,7\}$ denotes the index of the agent responsible for the posision of the end-effector. 

For agents who take control of the wrist joints, the observation can be written as :
\begin{equation}
\label{OBfun2}
    o_t^a=\langle p_b,\phi_b,v_b,\omega_b,q_\phi^a,\dot{q_\phi^a},\phi_e,\phi_d\rangle
\end{equation}
where $a=\{2,4,6,8
\}$ is the agent index, and the current orientation of the end-effector and the corresponding target attitude is denoted by $\phi_e$ and $\phi_d$ respectively. The three-dimensional output of the agent's policy determines the desired velocity of the three joints under its control.

\subsubsection{Reward Function}

For the trajectory planning task, in order to achieve the requirements of capture precision, we design a reward function including the distance error or attitude error between the end-effector and the target. Furthermore, to achieve smooth control and reduce energy consumption, the reward function is also dependent on the joint velocity at time $t-1$ and $t$. The following formula is the reward function: 
\begin{multline}
\label{r_catch}
r_t^a = -\big[w_1 \Vert e_*^a \Vert ^2 + \log(w_2 \Vert e_*^a \Vert ^2 + \epsilon) \\+ w_3 \Vert u^a_t - u^a_{t-1} \Vert ^2 + w_4 \Vert u^a_t \Vert ^2\big]
\end{multline}
The error $e$ is defined in Section \ref{pro_for} where the subscript $*$ denotes the position $p$ or orientation $\phi$. The middle part $\log(w_2 \Vert e_*^a \Vert ^2 + \epsilon)$ encourages smaller error $e$ to get higher reward, and the last two parts is designed to make the control process smooth and low energy consumption, respectively. In the experiments, we set $w_1 = 0.001, w_2 = 1, w_3 = 0.01, w_4 = 0.05$.

For the base orientation task, all the agents share the same reward, which can be written as: 
\begin{multline}
\label{r_base}
r_t = -\big[w_1 \Vert e_b \Vert ^2 + \log(w_2 \Vert e_b \Vert ^2 + \epsilon) \\+ w_3 \Vert u_t^a - u^a_{t-1} \Vert ^2 + w_4 \Vert u^a_t \Vert ^2 + w_5 co\big]
\end{multline}
where in the base reorientation task we need to ensure that the arms are collision-free with the base, so we add the term $w_5 \times co$, and $co$ equals to 1 if collision happens else 0. The parameter $w_5$ is set to 0.50, and the detailed value of the hyperparameters during training are illustrated in Table \ref{hp of algo}.

\begin{table}[t]
\begin{center}
\caption{Hyperparameters during Training}
\label{hp of algo}
\begin{tabular}{|c|c|c|}
\hline
Task& {\makecell[c]{Trajectory \\ Planning }} & {\makecell[c]{Base \\ Reorientation}} \\
\hline
Episode Length & 50 & 50 \\
Actor network & (512,512) & (512,512)\\
Critic network & (512,512) & (512,512)\\
Learning rate of actor & 8e-4 & 7e-4 \\
Learning rate of critic & 8e-4 & 7e-4 \\
Optimizer & Adam & Adam\\
Share Reward & False & True\\
Discount ($\gamma$)& 0.99 & 0.99\\
PPO clip ratio & 0.2 & 0.2 \\
PPO epoch & 5 & 5 \\
Entropy coefficient & 0.05 & 0.05 \\
Maximum steps & 20e6  & 20e6 \\
\hline 
\end{tabular}
\end{center}
\end{table}

\section{EXPERIMENTS}

In this section, we carry out a series of experiments to illustrate the precision, robustness and adaptability of the proposed multi-agent training paradigm. There are four objectives of the experiments: 
\begin{itemize}
    \item compare the performance of our multi-agent decentralized training paradigm with that of single-agent centralized training to verify the rationality of agent division methodology;
    \item conduct ablation studies, including comparison with other baseline algorithms; 
    \item evaluate the robustness of the learned strategies, including scenarios involving varying base mass, single arm failure and disturbance torques at the joints;
    \item recombine different strategies onto the same space robot so that it can adjust the posture of the base while completing grasping tasks.
\end{itemize}

\subsection{Comparison with Centralized Training}

\begin{figure}[t]
\centering
\includegraphics[width=0.48\textwidth]{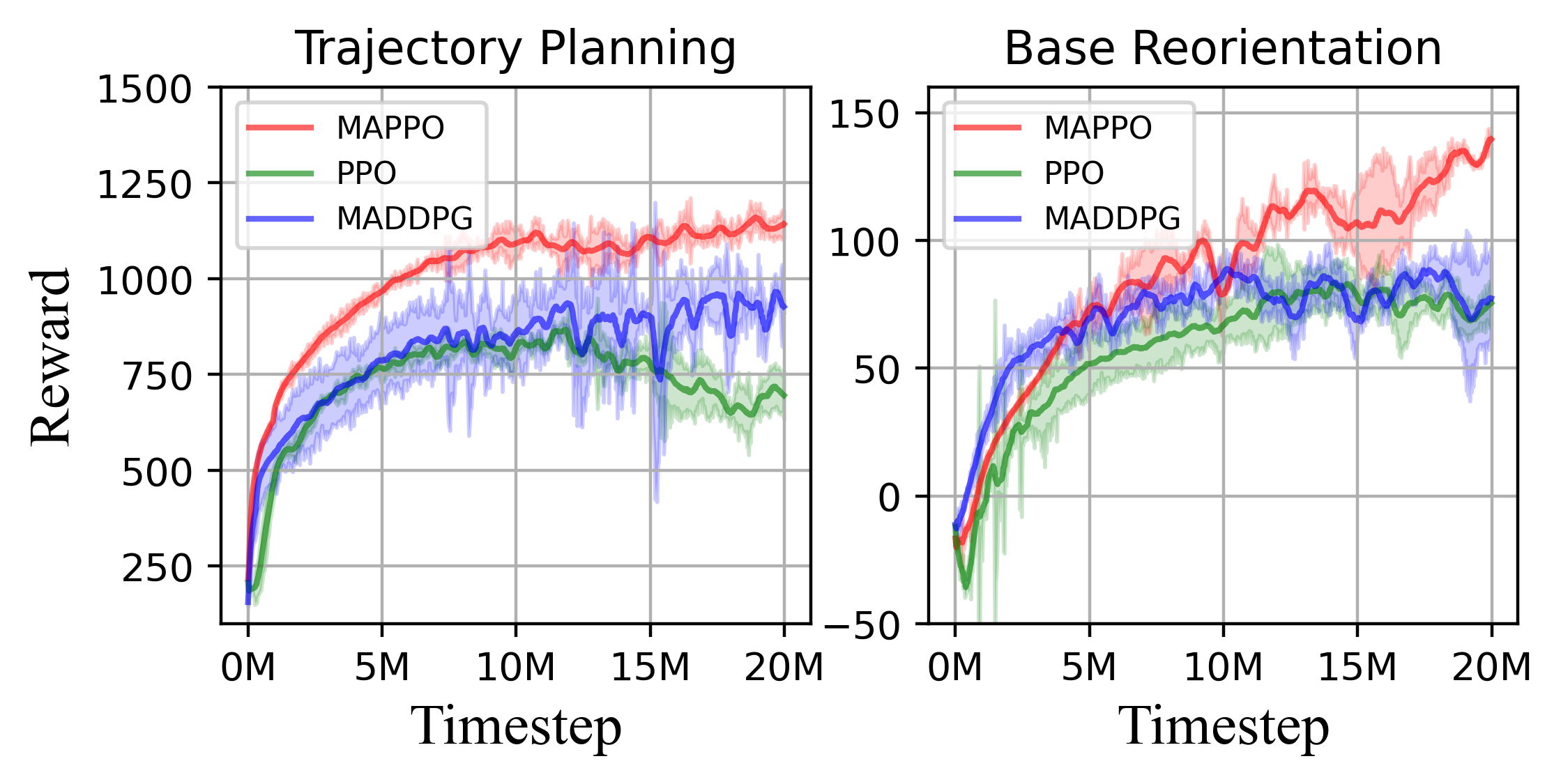}
\caption{Average performance for MAPPO, PPO and MADDPG over three seeds; the x-axis is training iteration. The rewards for all agents are added together in MAPPO and MADDPG in the trajectory planning task to make comparison with PPO. MAPPO outperforms the other two algorithms in both trajectory planning and base reorientation tasks. }
\label{cent_vs_dist}
\end{figure}

\begin{figure}[t]
\centering
\includegraphics[width=0.48\textwidth]{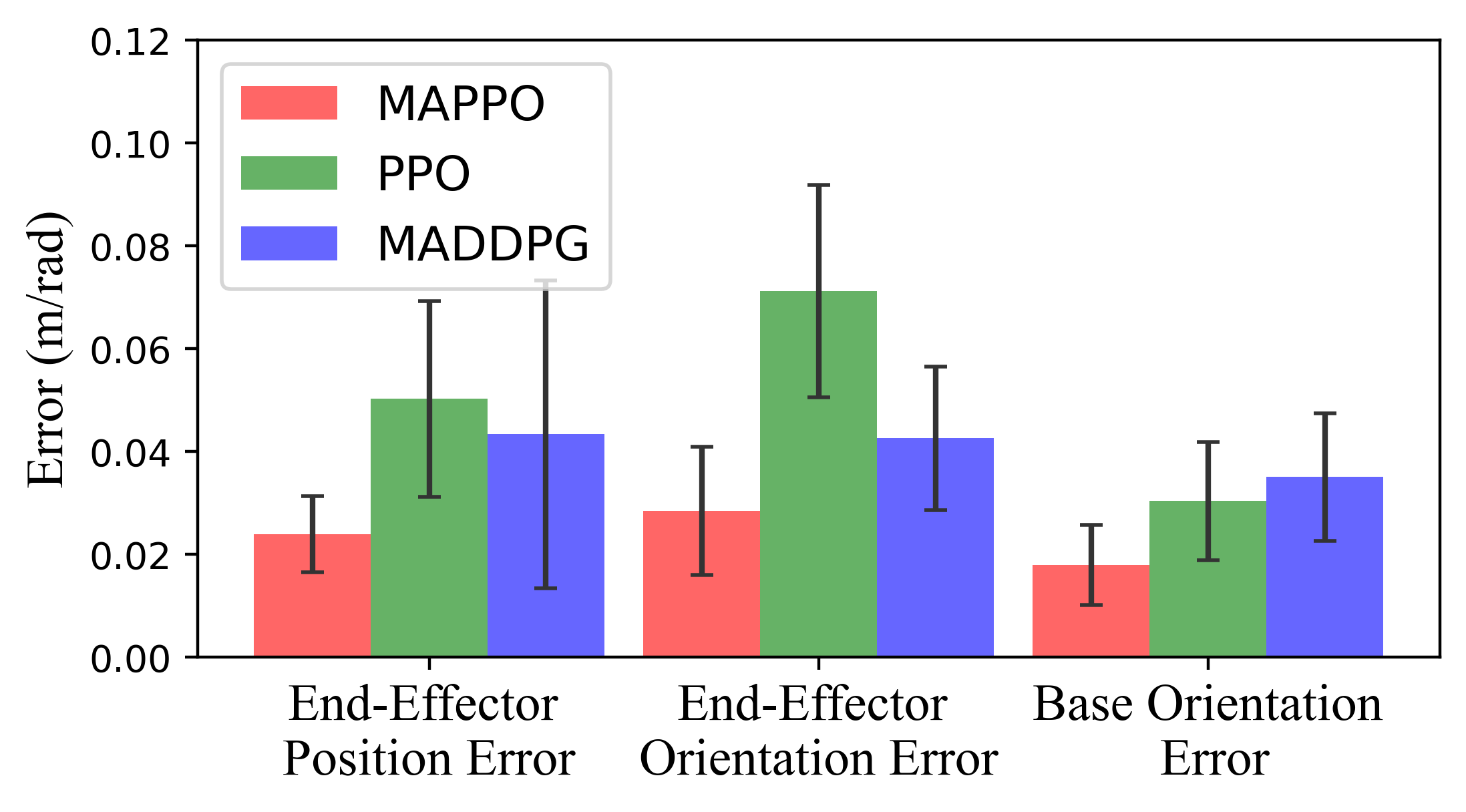}
\caption{The steady state error of the end-effector position and orientation and the base attitude of different algotithms. The results are obtained under 30 random seeds for each task.}
\label{comp_algo}
\end{figure}

To demonstrate the effectiveness of our proposed multi-agent decentralized training paradigm, we begin by comparing our algorithm with the single-agent RL algorithm which controls the arms in a centralized manner. 
The training curves of different algorithms, along with the corresponding test errors, and shown in Fig. \ref{cent_vs_dist} and Fig. \ref{comp_algo}, respectively. The results indicate that the training process becomes faster and more stable with higher rewards under the MARL paradigm. The mean position error of the end-effector is below 0.025 m with the mean orientation error below 0.04 rad (2.3$^\circ$) in the trajectory planning task, which is significantly less than that of the PPO algorithm (0.05 m / 0.08 rad). Similar improvements are observed in the base reorientation task. Furthermore, the smaller error variance of the MAPPO algorithm further demonstrates its superior generalization ability across different objectives. It can be inferred that the large exploration space due to the large number of joints to be controlled and the mixed single reward in the trajectory planning task increase the difficulty of exploration and learning. By adopting the MARL training paradigm, the control task is assigned to multiple agents, with each agent learns its own policy in a smaller exploration space, maximizes its individual reward or learns its own action's effect on the shared reward. 

\subsection{Ablation Experiments}

\begin{figure}[t]
    \centering
    \includegraphics[width=0.48\textwidth]{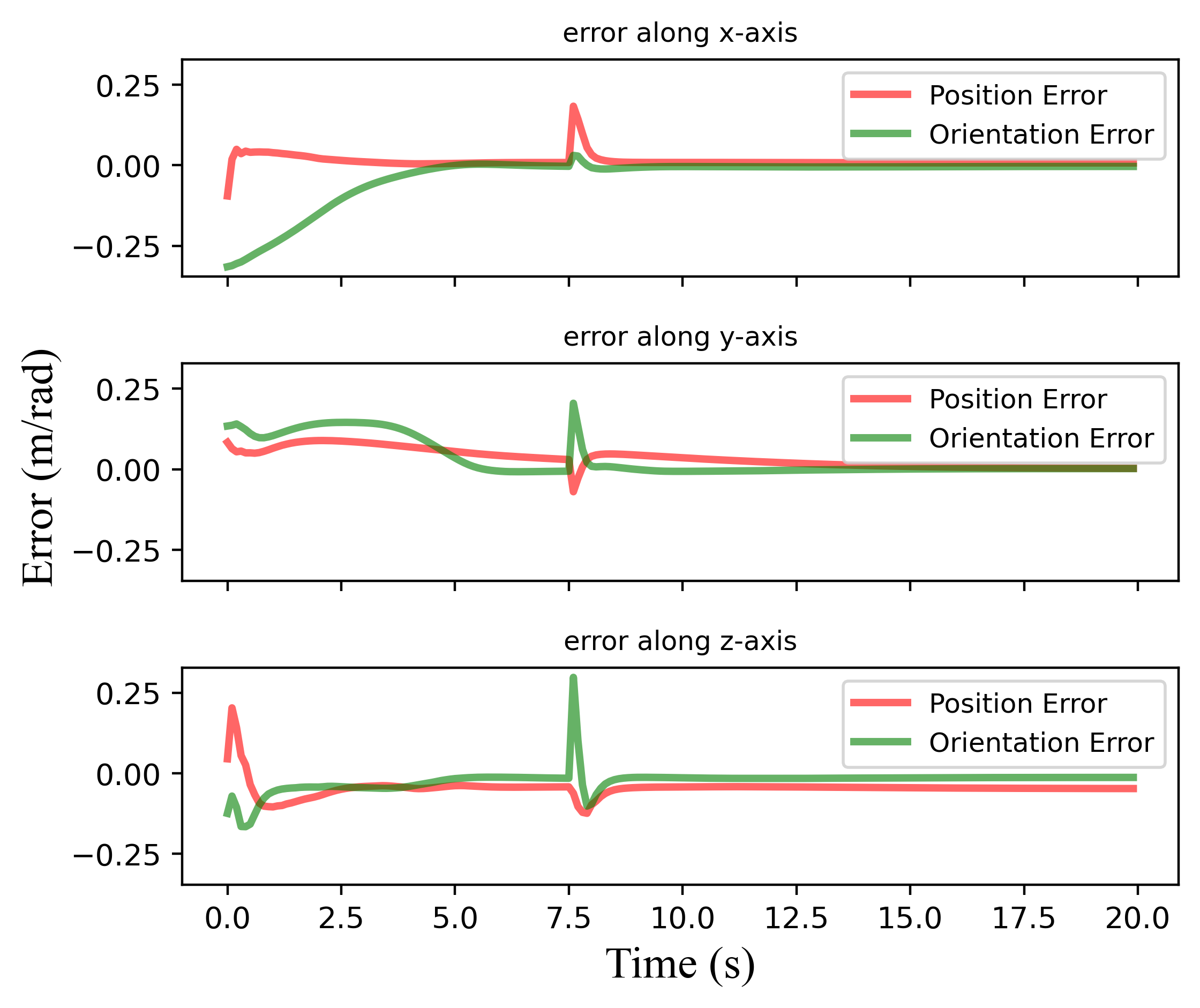}
    \caption{The error curves along the x,y and z axes in the presence of disturbance force in the trajectory planning task. The external force is exerted at 7.5s. }
    \label{4arm_dist}
\end{figure}

\begin{figure}[t]
    \centering
    \includegraphics[width=0.48\textwidth]{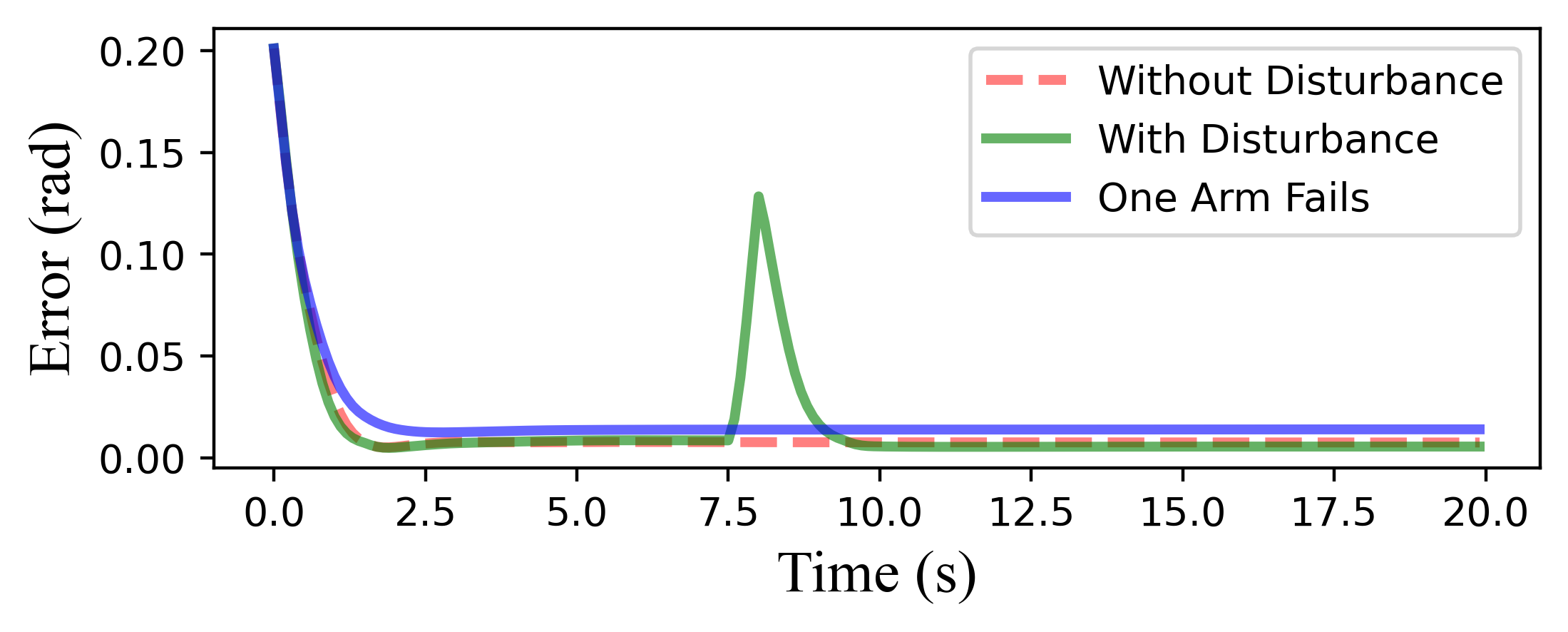}
    \caption{The error curves of the base in the base reorientation task under different scenarios. In the anti-disturbance experiment, the external force is exerted at 7.5s.}
    \label{base_error}
\end{figure}

In this section, we compare the performance of MAPPO algorithm with other baseline methods. Note that he continuous control feature and the unshared rewards make a large class of MARL algorithms inappropriate. Considering these constraints, we adopt MADDPG \cite{lowe2020multiagent} as the baseline methods. From Fig. \ref{cent_vs_dist} and Fig. \ref{comp_algo}, we can see that MADDPG achieves lower rewards and presision with higher variance in contrast with MAPPO. We hypothesize that the state-action value $Q^{\bm{\pi}}_a(s_t,\mathrm{\textbf{u}_t})$ learne by MADDPG has much higher input dimensions and is harder to learn its precise value compare to $V^\pi_a(s)$. Moreover, MAPPO is an on-policy RL algorithm, which improves the policy based on the latest policy and has been shown to be competitive and even better compared to off-policy methods including MADDPG while using a comparable number of samples\cite{yu2022surprising} . As a result, we adopt MAPPO instead of MADDPG as our training algorithm. 


\subsection{Evaluation of anti-disturbance ability}

\begin{figure}[t]
\centering
\includegraphics[width=0.48\textwidth]{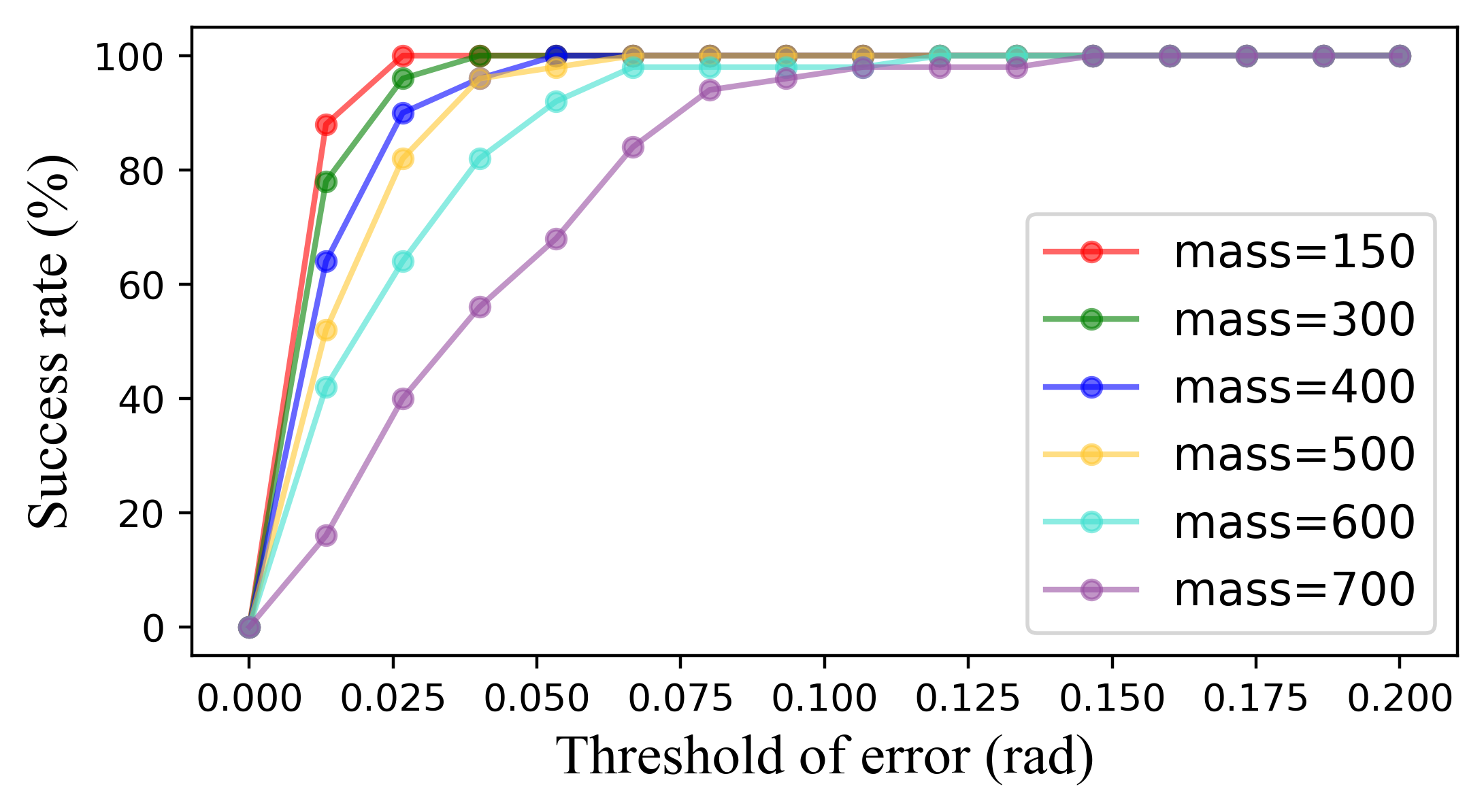}
\caption{The performance of the base reorientation task under different masses of the base. The policy is trained using a satellite with a mass of 400 kilograms.}
\label{diff_mass}
\end{figure}

\begin{figure}[t]

    \begin{subfigure}[t]{0.155\textwidth}
           \centering
           \includegraphics[width=\textwidth]{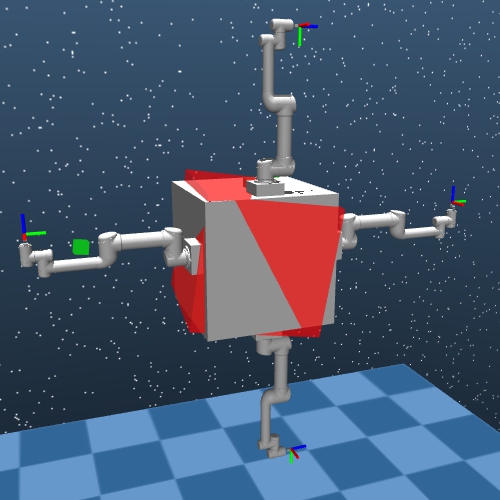}
            \caption{}
    \end{subfigure}
    \begin{subfigure}[t]{0.155\textwidth}
            \centering
            \includegraphics[width=\textwidth]{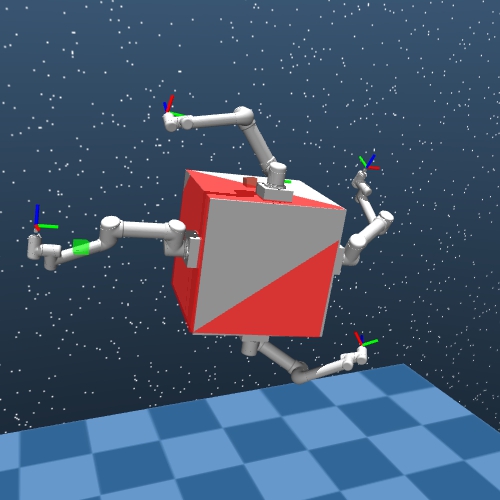}
            \caption{}
    \end{subfigure}
    \begin{subfigure}[t]{0.155\textwidth}
           \centering
           \includegraphics[width=\textwidth]{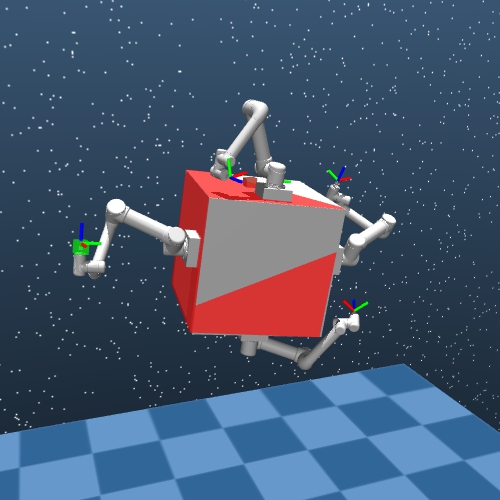}
            \caption{}
        \end{subfigure}
    \begin{subfigure}[t]{0.48\textwidth}
           \centering
           \includegraphics[width=\textwidth]{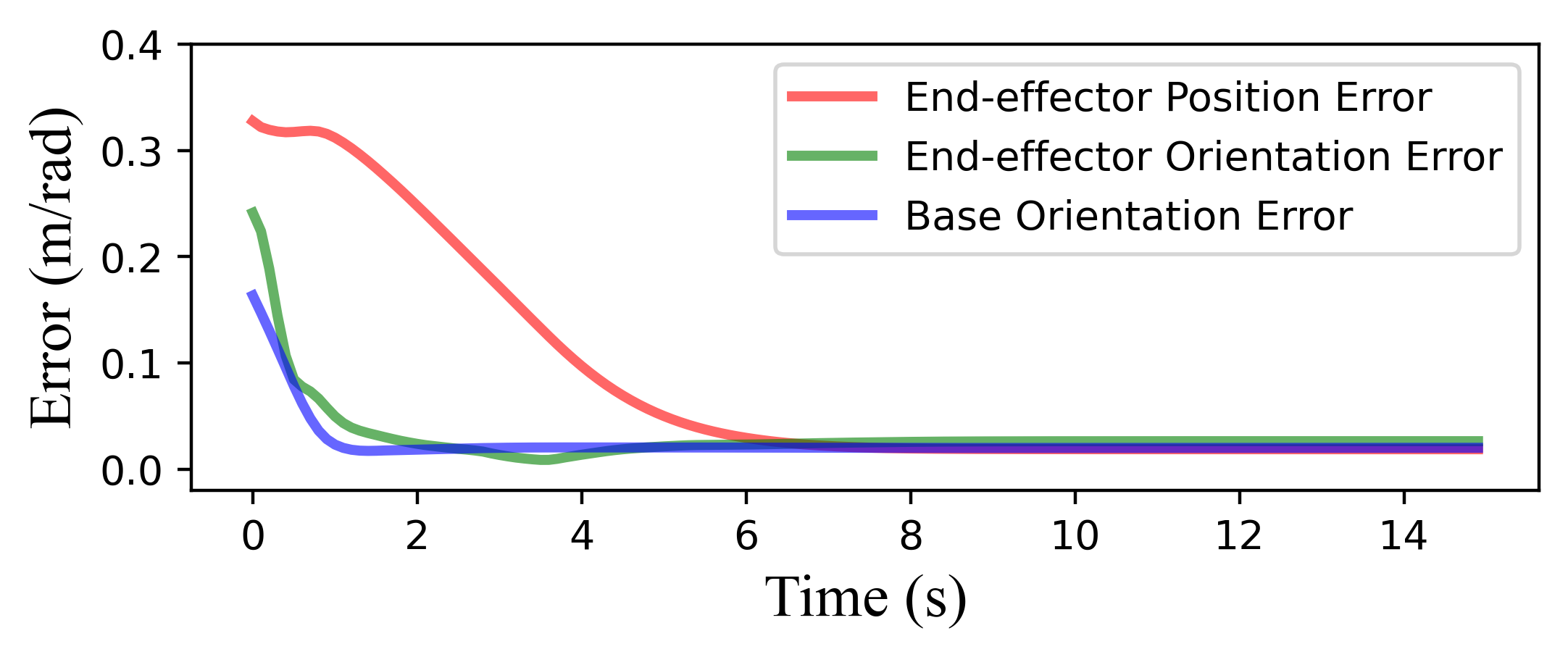}
            \caption{}
    \end{subfigure}
    \caption{(a) - (c) The adjustment process of the space robot with reassembled policies. The left arm aims to reach the desired position and orientation in green, while the other three arms work on adjusting the base's attitude. (d) The errors of the end-effector and base attitude during adjustment.}
    \label{mixed_goal}
\end{figure}

To assess the robustness and anti-disturbance ability of the learned policy, we introduce disturbance torques at the joints to test the policy's resilience. 
Specifically, in the trajectory planning task we apply an external force to the end-effector at 7.5 s. The position and orientation errors along the x, y and z axes are shown in Fig. \ref{4arm_dist}. It is observed that the errors along all axes converge to below 0.015 m / 0.02 rad, demonstrating the learned policy's superior robustness and its ability to replan the trajectory when faced with disturbances. In the base reorientation task, we additionally evaluate the policy's performance when one arm fails, with the error curves displayed in Fig. \ref{base_error}. The errors converge to below 0.03 rad regardless of the scenario, indicating that the trained policy possesses significant anti-disturbance capabilities and remains robust even with the failure of one robotic arm. Furthermore, we change the mass of the base satellite and check the performance of the base reorientation task. The success rates under different criterions are depicted in Fig. \ref{diff_mass}, illustrating that the trained policies are still effective with different satellite masses, achieving a 90\% success rate in most cases under the threshold error of 0.05 radians. However, when there is a substantial change in mass, additional training may be necessary to enable the agent to learn how to adjust the posture of a heavier base.

\subsection{Reassembly of policies}

After completing the training for trajectory planning and base reorientation tasks, we sought to determine whether these two strategies could be recombined, thereby enabling some robotic arms to execute the trajectory planning task while others adjust the posture of the base, akin to an octopus during hunting. Note that centralized control would require retraining a new policy to accomplish this mixed task. However, the MARL training paradigm allows us to take advantage of the flexibility of decentralized training properties since each agent maintains its individual policy and can be replaced as needed. To verify the feasibility of the approach, we reassemble the previously trained strategies, assigning the base reorientation strategy to three robotic arms and the trajectory planning task to the remaining one. The adjustment process and the errors of the end-effectot and base attitude are illustrated in Fig. \ref{mixed_goal}. The results indicate that through strategy restructuring, space robots can use some arms to maintain or adjust the base posture while others execute trajectory planning, with the end-effector position error below 0.04 m, orientation error below 0.045 rad, and the base orientation error below 0.035 rad. This outcome lends significant value to our research, demonstrating that the training paradigm of multi-agent systems allows for the flexible combination of different strategies to accomplish specific composite tasks without the necessity of retraining.

\section{CONCLUSIONS}

This paper proposes a decentralized multi-agent reinforcement learning paradigm for trajectory planning and base reorientation tasks for multi-arm space robots. Inspired by the hunting behaviors of octopuses, we move away from the traditional centralized control strategy and hierarchically assign the control task of the robotic arms to different agents across three levels. The ablation studies demonstrate that, with a smaller exploration space and clearer reward signals, the individual agents can more easily and effectively find the control policies. We demonstrate the robustness of the learned policies under standard conditions, including the joint disturbances, varying masses of the base satellite, and the failure of one arm. Furthermore, by leveraging the flexibility of the multi-agent training paradigm, we can reassemble various policies trained for different tasks, enabling the space robot to execute trajectory planning with certain arms while reorienting its base with the others. 

\bibliographystyle{unsrt}  

\bibliography{main}

\end{document}